\documentclass{article}

\usepackage{microtype}
\usepackage{graphicx}
\usepackage{subfigure}
\usepackage{booktabs} 
\usepackage{amsmath}
\usepackage{amssymb}
\usepackage{amsfonts}

\usepackage{hyperref}

\usepackage[accepted]{icml2020}

\icmltitlerunning{Unsupervised Domain Adaptation in the Absence of Source Data}

\begin{document}

\twocolumn[
\icmltitle{Unsupervised Domain Adaptation in the Absence of Source Data}



\icmlsetsymbol{equal}{*}

\begin{icmlauthorlist}
\icmlauthor{Roshni Sahoo}{to,equal}
\icmlauthor{Divya Shanmugam}{to,equal}
\icmlauthor{John Guttag}{to}
\end{icmlauthorlist}
\icmlaffiliation{to}{CSAIL, MIT}

\icmlcorrespondingauthor{Roshni Sahoo}{rsahoo@mit.edu}
\icmlcorrespondingauthor{Divya Shanmugam}{divyas@mit.edu}

\icmlkeywords{Distribution Shift, Unsupervised Domain Adaptation}

\vskip 0.3in
]



\printAffiliationsAndNotice{\icmlEqualContribution} 

\begin{abstract}
 Current unsupervised domain adaptation methods can address many types of distribution shift, but they assume data from the source domain is freely available. As the use of pre-trained models becomes more prevalent, it is reasonable to assume that source data is unavailable. We propose an unsupervised method for adapting a source classifier to a target domain that varies from the source domain along natural axes, such as brightness and contrast. Our method only requires access to unlabeled target instances and the source classifier. We validate our method in scenarios where the distribution shift involves brightness, contrast, and rotation and show that it outperforms fine-tuning baselines in scenarios with limited labeled data.
\end{abstract}

\section{Introduction}
Machine learning methods operate under the assumption that training data and test data come from the same distribution. When
this assumption is violated, the performance of a model trained on the source domain (training distribution) will degrade when tested on the target domain (test distribution) \cite{robustness}. This problem is widespread; sensitivity to test-time perturbations has been shown in facial recognition software \cite{face}, medical image analysis \cite{bias_field}, and self-driving car vision modules \cite{selfdriving}. 

Domain adaptation techniques rely on labeled source examples and no (or few) labeled target examples to address this problem \cite{recent_survey}. In unsupervised domain adaptation (UDA), no labeled target examples are available. Previous works in UDA fall into two main classes. The first class aims to align representations of the source and target domains in some feature space \cite{ganin}. The second class of methods uses generative models to transform source images to resemble target images \cite{pixelgan}. 

These methods can address general forms of distribution shift but remain limited by the assumption of freely available source data. The source data may be inaccessible, for example, due to contractual obligations between data owners and data customers \cite{domain_adaptation_no_source}. In addition, as the usage of pre-trained models rises in popularity, it is common to have access to a model but not the data on which it was trained. 

\begin{figure}[t]
    \centering
    \includegraphics[height=0.15\textheight]{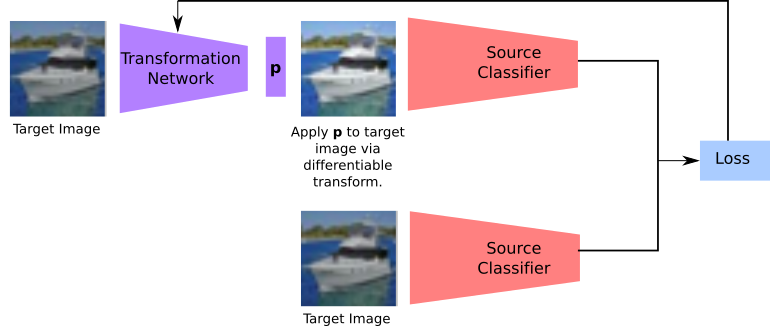}
    \caption{Training pipeline. The target image is passed to the transformation network, which predicts transformation parameters $\hat{p}$. The transformation function $g_{\hat{p}}$ is applied to the image. Next, the loss is computed, rewarding parameters that result in a high maximum softmax probability with the source classifier $\phi$.}
    \label{fig:method}
\end{figure}

With stricter assumptions on the nature of the distribution shift, we propose a method for unsupervised domain adaptation in the absence of source data. We consider settings in which the target domain is shifted from the source along natural axes of variation. A realistic use case is adapting classifiers trained on medical images. Differences in protocols can cause variation in resolution, intensity profile, and contrast for MRI volumes  \cite{nature_medical} and chest X-rays \cite{lenga2020continual}. Furthermore, chest X-rays may suffer from geometric deformations resulting from poor scan conditions, and aligning the images can improve performance on a downstream classification task \cite{align_chest}.

Our method leverages the softmax probabilities of the source classifier to learn transformations that bring target images closer to the source domain (Figure \ref{fig:method}). In our evaluations, we demonstrate that learning transformations can recover accuracy lost by the source classifier on various target domains. Furthermore, we find that our unsupervised method outperforms fine-tuning in label-scarce settings.

\section{Problem Definition}


We aim to produce transforms that map images in the target domain to the source domain, with the goal of improving accuracy on a $C$-way classification task.

\textbf{Definitions.} A domain consists of an image space $\mathcal{X}$ and a label space $\mathcal{Y}.$  The source domain is ${\mathcal{X}^s \times \mathcal{Y}^{s}}$, and  the target domain is ${\mathcal{X}^{t} \times \mathcal{Y}^{t}}$. Both $\mathcal{X}^{s}, \mathcal{X}^{t} \subset \mathcal{X}$ and $\mathcal{Y}^{s}, \mathcal{Y}^{t} \subset \mathcal{Y}$. A transform is a function $f_{p}: \mathcal{X} \rightarrow \mathcal{X}$, where $p \in \mathcal{P}_f,$ the space of transformation parameters.

\textbf{Distribution Shift.} We model the distribution shift from source to target domain as a non-deterministic application of label-preserving forward transforms $\mathcal{F} = \{f_{p} | p \in \mathcal{P}_{f}\}$. For a fixed $\mu$ and $\sigma$, we can generate
\[\mathcal{X}^{t} = \{f_{p_{i}}(x_{i}) | p_{i} \sim N(\mu, \sigma), x_{i} \in \mathcal{X}^{s}\}.\]
To generate each image in $\mathcal{X}^{t}$, a new forward transform $f_{p_{i}}$ is sampled and applied to a source image $x_{i}$. We restrict the choice of $\mu$ and $\sigma$ such that $p_{i} \in \mathcal{P}_{f}$ with high probability.

\textbf{Assumptions.} To adapt to this distribution shift, we require the following inputs:
\begin{enumerate}\itemsep0em
\item A source classifier $\phi : \mathcal{X} \rightarrow \mathbb{R}^C$. The classifier's training set is sampled from $\mathcal{X}^{s} \times \mathcal{Y}^{s}.$ The classifier $\phi$ produces class probabilities through a softmax layer. In practice, $\phi$ can be a pre-trained neural network.
  \item A set of $N$ images $\mathbf{T} = \{x'_i\}_{i=1}^N$, where $x'_i \in \mathcal{X}^t$.
 \item A class of differentiable backward transforms, ${\mathcal{G} = \{g_{p}| p \in \mathcal{P}_{g} \}}$. Transforms in $\mathcal{G}$ are applied to examples from $\mathcal{X}^t$. 
\end{enumerate}
We assume no access to data from the source domain (${\mathcal{X}^{s}  \times \mathcal{Y}^{s}}$) or labels from the target domain ($\mathcal{Y}^{t}$). 

\textbf{Learning a Transformation Network.} Our goal is to recover $\mathcal{X}^{s}$ by learning the optimal transform $g_{p} \in \mathcal{G}$ for each target image in $\mathbf{T}$. If $\mathcal{F}$ consists of invertible functions and $\mathcal{G}$ contains the inverses of functions in $\mathcal{F}$, one can recover $\mathcal{X}^s$. If not, one can approximate $\mathcal{X}^{s}$. Learning the transformation parameter(s) is useful because we can apply the model $\phi$ to the transformed test set and achieve improved accuracy compared to running inference on $\mathbf{T}$ directly. The difficulty of this task depends on the shift severity and shift range, which are represented by $\mu$ and $\sigma$ respectively.

\section{Method}

Our method consists of two steps: 1) learning transformation parameters that bring the target images closer to the source domain, and 2) transforming the target examples with the learned parameters and running inference on the resulting images using the source classifier $\phi$.

Previous work in out-of-distribution detection demonstrates that in-distribution examples tend to have greater maximum softmax probabilities (MSP) than out-of-distribution examples \cite{hendrycks}. In addition, temperature scaling, a calibration procedure where the outputs of a classifier are scaled prior to applying the softmax layer, further enlarges the MSP gap between in-distribution and out-of-distribution examples \cite{ood_liang}.

Under distribution shift, we expect most $x \in \mathbf{T}$ to be out-of-distribution for $\phi$. As a result, we develop a loss function that rewards predicted parameters that maximize the temperature-scaled MSP of the transformed image  relative to that of the original image. Let $s$ be the temperature-scaling constant. Given an image $x$, we aim to maximize the MSP gap between the transformed image and the original image as follows
\[\max_{p \in \mathcal{P}_{g}} \text{  MSP}(\phi(g_{p}(x))/s) -\text{MSP}(\phi(x)/s).\]

We aim to predict the optimal $p$ for each image by training a transformation network. The transformation network maps target examples to transformation parameters $\hat{p}$. To train the network, we minimize the following loss function
\[\mathcal{L}(\hat{p}; x ) = - \Big(\text{MSP}(\phi(g_{\hat{p}}(x))/s) -\text{MSP}(\phi(x)/s)\Big). \]
The second term of the loss is a constant with respect to $\hat{p}$, so it does not affect the optimization of the transformation network, but we include it so that the converged loss value is a meaningful quantity (a proxy for the distance between the transformed images and the original target images).

With the trained network, we predict $\hat{p}$ for each image, apply the transformations to the corresponding images, and run inference on them using $\phi$.

\textbf{Implementation.} We clamp the outputs of the transformation network so that all predicted parameter values are constrained to $\mathcal{P}_{g}$. We initialize the bias parameters of the network's last layer with $p$ such that $g_{p}(x) = x.$  The temperature scaling constant $s$ is typically chosen using a validation set \cite{temp_scaling}, but \citeauthor{ood_liang} show that simply using a large constant $s$ is sufficient. In our method, we use $s=10.$ Architecture and training details are provided in Section \ref{sec:tnet}.

\section{Experiments}
First, we investigate the trade-off between fine-tuning, an adaptation technique that requires labeled target examples, and our unsupervised method. We find that we outperform fine-tuning in label-scarce settings. Second, we evaluate our method's sensitivity to the severity and range of the distribution shift and show that the proposed method can achieve accuracies on par with a classifier trained on the target domain. In this section, we show results on CIFAR-10, where the source classifier is a ResNet-18 model. Further CIFAR-10 experiments (on fine-tuning and coping with shifts along multiple axes of variation) and MNIST experiments corroborate these results and can be found in the supplement, along with details on the experiment setup.

\begin{subsection}{Setup Overview}
\label{sec:setup}

\textbf{Distribution Shift.} We use the following forward transforms:
\begin{itemize} \itemsep0em
	\item $b_{p}(x)$: Scales the brightness of an image $x$ by a factor of $p$, where $p \in  \mathcal{P}_{b} = [0, \infty).$
	\item $r_{p}(x)$: Rotates an image $x$ by $p$ degrees, where $p \in \mathcal{P}_{r} = [-180, 180].$
	\item $c_{p}(x)$: Scales the contrast of an image $x$ by a factor of $p$, where $p  \in \mathcal{P}_{c} = [0, \infty).$
\end{itemize}

To simulate distribution shift, we use one or more forward transforms from above. For each selected transform, we pick a corresponding $\mu$ and $\sigma$, which govern the distribution of forward transform parameters. We express the forward transforms for a brightness shift as
\[B_{\mu, \sigma} = \{b_{p_{i}}| p_{i} \sim N(\mu, \sigma) \}.\]

Overloading the notation, we apply $B_{\mu, \sigma}$ to a dataset ${D = \{(x_{i}, y_{i})\}}$ as follows
\[B_{\mu, \sigma}(D) = \{(b_{p_{i}}(x_{i}), y_{i}) | (x_{i}, y_{i}) \in D\}.\]

We use the same notation for contrast ($C_{\mu, \sigma}$) and rotation ($R_{\mu, \sigma}$) shifts, as well.

\textbf{Backward Transforms.}  For all experiments, we assume that the distribution shift occurs along the axes of rotation, brightness, and contrast. Accordingly, we set the class of backward transforms to be
\[ \mathcal{G} = \{r_{p_{1}} \circ b_{p_{2}} \circ c_{p_{3}} | p_{1}  \in \mathcal{P}_{r}, p_{2} \in \mathcal{P}_{b}, p_{3} \in \mathcal{P}_{c} \}.\]

\textbf{Datasets.} We consider the CIFAR-10 dataset and use the pre-processing pipeline from the PyTorch model zoo \cite{NEURIPS2019_9015}. We construct target domains by varying the contrast and brightness of the CIFAR-10 test set. We select these target domains because contrast and brightness changes are common corruptions on natural images. We do not evaluate adaptation to rotation shift on natural images because there is an artificial correlation between the optimal transformation and the size of the black artifacts at the corners of the rotated image.

\textbf{Baselines.} For each target domain, we assess performance of two baselines. The first baseline is the source classifier trained on the CIFAR-10 training set. The second baseline is an oracle model trained on the target domain. We expect the oracle model to outperform our method because it is tested on the domain on which it is trained. To generate the training dataset for the oracle model, we apply forward transforms to the CIFAR-10 training set. The source classifier and oracle model have the same architecture. 

\end{subsection}

\begin{subsection}{Comparison to Fine-Tuning}
\label{sec:fine_tuning}
\begin{figure*}[t]
\centering
    \includegraphics[width=\textwidth]{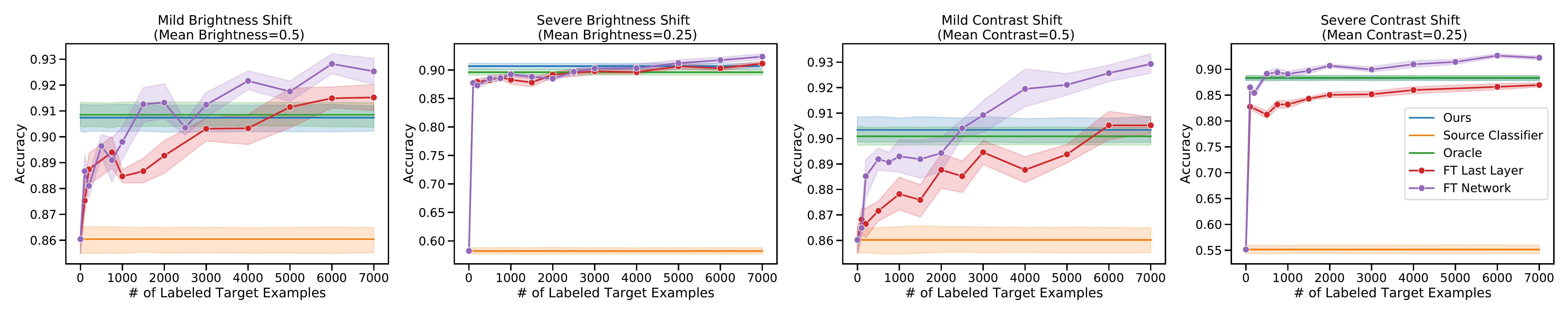}
    \caption{Accuracy achieved by fine-tuning a source classifier with $n$ annotated target examples. Our unsupervised method outperforms fine-tuning the last layer (FT Last Layer) when there are less than $4000$ labeled target examples and fine-tuning the entire source classifier (FT Network) when there are less $1000$ labeled target examples. }
    
    \label{fig:tl} 
\end{figure*}

Fine-tuning is a common technique for adapting a source classifier to a target domain in the presence of labeled target examples \cite{fine_tuning}. We compare our method, which does not use any labeled target examples, to fine-tuning the source classifier on $n$ labeled target examples.

\textbf{Baselines.} In addition to the baselines in Section \ref{sec:setup}, we compare our method to two fine-tuning schemes. In both schemes, a ResNet-18 model is initialized with the source classifier's weights and is trained on labeled target examples. In the first scheme, we fine-tune the last layer, freezing all other model weights. In the second scheme, we fine-tune the entire network, permitting all model weights to be updated. 

\textbf{Evaluation.} We consider mild and severe shifts along the axes of brightness and contrast. For mild shifts, the target domains are generated by applying the forward transforms $B_{0.5, 0.05}$ and $C_{0.5, 0.05}$ to the CIFAR-10 test set. For severe shifts, we apply $B_{0.25, 0.05}$ and $C_{0.25, 0.05}$. These target domains are low brightness and low contrast settings; experiments on high brightness and high contrast target domains are in Section \ref{sec:supp_fine_tuning}.

We evaluate each method on $30\%$ of the examples from the target domain and produce error bars through repeated subsampling. Our unsupervised method is trained on images from the remaining $70 \%$ of the target domain. The fine-tuning baselines are trained on $n$ labeled examples from the same 70\% of the target domain. Of the $n$ labeled examples, one-fifth are used for validation. Note that in real-world deployment of our unsupervised method, an entire unlabeled test set can be used for both training and inference.

\textbf{Results.} Across these shifts, our method, which uses $0$ labeled target examples, outperforms fine-tuning the last layer of the source classifier when there are less than $4000$ labeled target examples and fine-tuning the entire network when there are less than $1000$ labeled target examples (Figure \ref{fig:tl}). As the number of labeled target examples increases, the fine-tuning methods improve accuracy on the target domain. The fine-tuning approaches and our method improve model performance relative to the source classifier.

\end{subsection}

\begin{subsection}{Effect of Shift Severity and Range}
\label{sec:severity_range}

\begin{figure*}[t]
\centering
    \includegraphics[width=\textwidth]{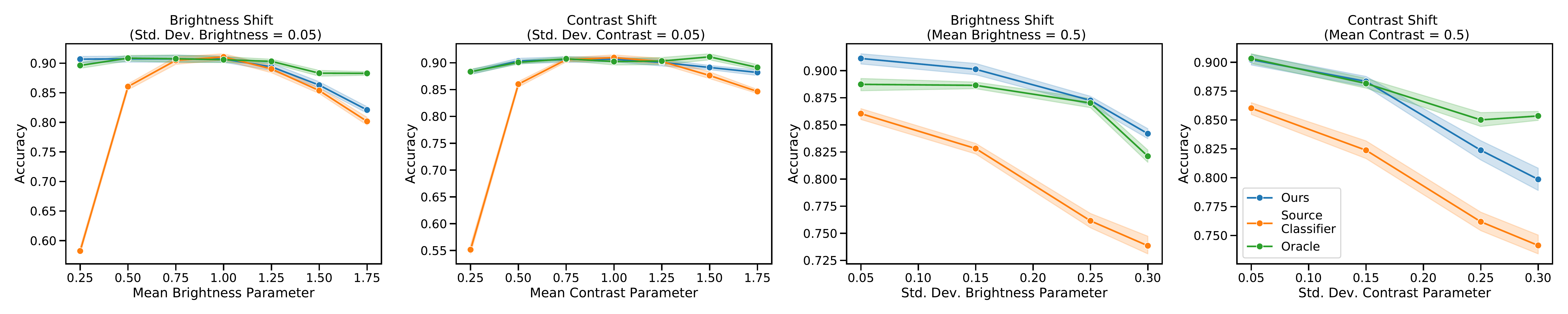}
    \caption{Performance of our method as the distribution shift varies in severity (leftmost plots) and in range (rightmost plots).}
    \label{fig:dist}
\end{figure*}

\textbf{Baselines.} We compare our method to the two baselines described in Section \ref{sec:setup}. 

\textbf{Evaluation.} We capture the shift severity and range with $\mu$ and $\sigma$, the mean and standard deviation of the forward transform parameters applied to generate the target domain. Let $D$ be the original CIFAR-10 test set. For the shift severity experiments, we assess performance on the following target domains
\[\{\mathcal{D}^{t}\} = \{B_{\mu, 0.05}(D) | \mu \in [0.25, 0.50 \dots 1.75] \}.\] For the shift range experiments, we assess performance on the following target domains
\[ \{\mathcal{D}^{t} \} = \{ B_{0.5, \sigma}(D) | \sigma \in [0.05, 0.15, \dots 0.30] \}.\] 
In the shift range experiments, we set $\mu=0.5$ to simulate a mild distribution shift (in contrast, with $\mu=1$, most generated examples are still in-distribution). The target domains generated by contrast shifts can be defined analogously. We evaluate the methods on random subsamples of $30\%$ of the target domain. Images from the remaining $70\%$ are used to train our unsupervised method.

\textbf{Shift Severity Results.} While the performance of the source classifier declines as $\mu$ moves further from the default setting of $\mu=1$ (Figure \ref{fig:dist}, leftmost plots), our method is often able to recover the lost accuracy. Our method achieves similar accuracy to the oracle model for all contrast shifts and for brightness shifts where $\mu \leq 1.$ Although it performs better than the source classifier for brightness shifts where $\mu >1$, our method does not match the accuracy of the oracle model in this range. Our method is limited by how well it can reverse the effect of the forward transform; in this case, we cannot easily add color back to an overexposed image.

\textbf{Shift Range Results.} As shift range increases, we observe that all methods decline in accuracy (Figure \ref{fig:dist}, rightmost plots). As the brightness shift range increases, the accuracy of our method declines more gradually than that of the source classifier. As the contrast shift range increases, both decrease in accuracy at a similar rate.

\end{subsection}



\begin{section}{Conclusion}
In contrast to previous UDA methods which rely on source data, we demonstrate that unlabeled data from a target domain and a source classifier can be leveraged to adapt to distribution shift along natural axes. This work may have applications in medical imaging, where target domain annotations are costly and data from the source domain is confidential and unavailable. Our future work includes extending our method to cope with more corruptions suggested by \citeauthor{robustness} and using our method to cope with bias field corruption in MRI images \cite{bias_field}.
\end{section}

\bibliography{ref}
\bibliographystyle{icml2020}

\section{Supplementary Material}

\begin{subsection}{Related Work}
We give an overview of domain adaptation in the absence of source data. Our method draws inspiration from previous works in the out-of-distribution detection literature, so we outline the parallels.

\textbf{Domain Adaptation without Source Data.} In this setting, practitioners lack access to any data from the source domain. The lack of source data in this task distinguishes it from the majority of existing domain adaptation methods. With no or limited access to source data, existing domain adaptation methods consider the source classifier's decisions as augmented features of the target data \cite{domain_adaptation_no_source}. Our approach similarly leverages the source classifier's predicted class probabilities to perform adaptation. 

Our method is also \textit{unsupervised}, which further assumes that we do not have access to labeled instances of the target domain. Work by \citeauthor{adscm} trains marginalized stacked denoising autoencoders (mSDA) on the unlabeled target data and aggregations of the source data. Although this approach has low computational cost, it is not directly applicable to visual data (images). Since the method operates on feature vectors, either the images must be flattened into vectors or embeddings of images must be generated. After that, the adaptation is performed on the vectors.


Our work is most closely related to a method that operates by updating model parameters using a self-supervised loss during test-time \cite{sun2019test}. While the method does not require source data, it assumes the inclusion of the self-supervised loss in the original model’s training regime. Our approach does not modify the original training process and thus, can be applied to pre-trained models.



\textbf{Out-of-Distribution Detection} Many works in out-of-distribution detection use softmax probabilities of classification models to determine whether an input is out-of-distribution \cite{hendrycks, ood_liang}. Although the prediction probability from a softmax distribution has a poor direct correspondence to confidence \cite{evidential}, \citeauthor{hendrycks} demonstrate that the maximum softmax probability (MSP) of out-of-distribution examples tends to be lower than the MSP of in-distribution examples, so MSP statistics are often sufficient for detecting whether an example is abnormal. 

Our method is similar to these out-of-distribution detection works because we use the MSP to quantify whether a transformation brings target instances closer to the source domain. This practice is common---out-of-distribution metrics are the foundation of multiple methods to address domain shift \cite{volpi2018generalizing, karani2020test}. We solve a different problem; we are interested in correcting out-of-distribution inputs.

\end{subsection}

\begin{subsection}{Additional CIFAR-10 Results}

\begin{subsubsection}{Comparison to Fine-Tuning}
\begin{figure*}
\centering
    \includegraphics[width=\textwidth]{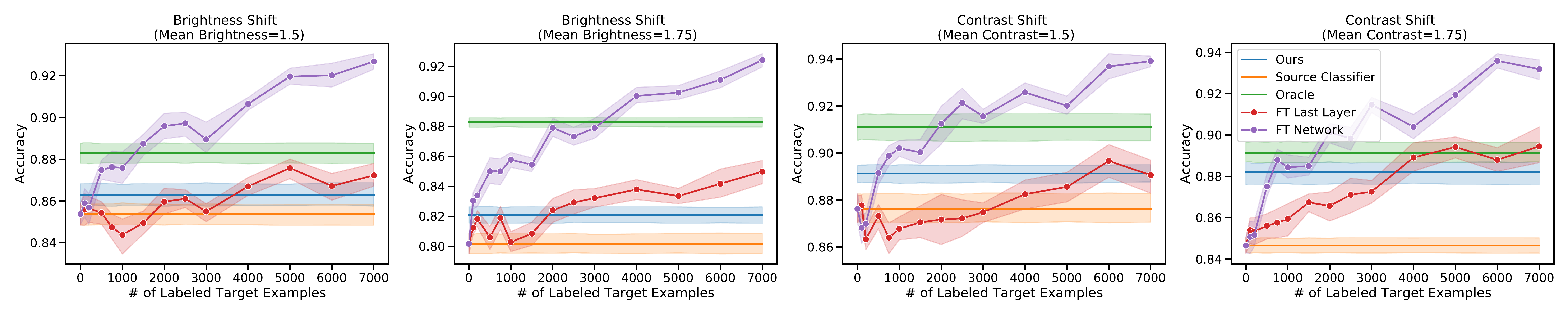}
    \caption{The target domains represent modest overexposures and contrast shifts. Fine-tuning (the entire network or only the last layer) on too few examples in the case of subtle shifts, $B_{1.50, 0.05}$ and $C_{1.50,0.05}$, degrades model accuracies below the performance of the original classifier, suggesting that our method may be a useful alternative in label-scarce settings.}
    \label{fig:modest_shifts}
\end{figure*}

\label{sec:supp_fine_tuning}
Continuing the experiments of Section \ref{sec:fine_tuning}, we compare the performance of fine-tuning to our method on high brightness and high contrast target domains.

\textbf{Baselines.} We use the same baselines as described in Section \ref{sec:fine_tuning}.

\textbf{Evaluation.} Brightness transforms with ${\mu >1}$ may not be label-preserving because excessive overexposure will remove relevant information for classification, so we consider modest overexposures. We evaluate on target domains generated by applying the forward transforms $B_{1.50, 0.05}$, $B_{1.75, 0.05}$, $C_{1.50, 0.05},$ and $ C_{1.75, 0.05}$ to the CIFAR-10 test set. Otherwise, our evaluation method is the same as in Section \ref{sec:fine_tuning}. 

\textbf{Results.} We observe that our method provides small improvements over the source classifier in these experiments (Figure \ref{fig:modest_shifts}). These improvements are not as pronounced as in Section \ref{sec:fine_tuning}. We hypothesize that this is because the target and source domains are more similar, so it is more difficult to detect whether examples are out-of-distribution. In addition, as mentioned in Section $\ref{sec:severity_range}$, the brightness transform is not invertible for ${\mu > 1}$, so our method can at best approximate the source domain and does not match the performance of the oracle model. At the same time, fine-tuning (the entire network or only the last layer) on too few examples in the case of subtle shifts $B_{1.50, 0.05}$ and $C_{1.50,0.05}$ results in lower accuracy than the original source classifier (Figure \ref{fig:modest_shifts}- leftmost and middle right plot). This suggests that our method may be a useful alternative for coping with slight perturbations in settings where there are fewer than $500$ labeled examples. 
\end{subsubsection}

\begin{subsubsection}{Effect of Shifts Along Multiple Axes}

\textbf{Baselines.} We use the same baselines as in Section \ref{sec:setup}, the source classifier and oracle models.

\begin{figure*}
\centering
    \includegraphics[width=0.5\textwidth]{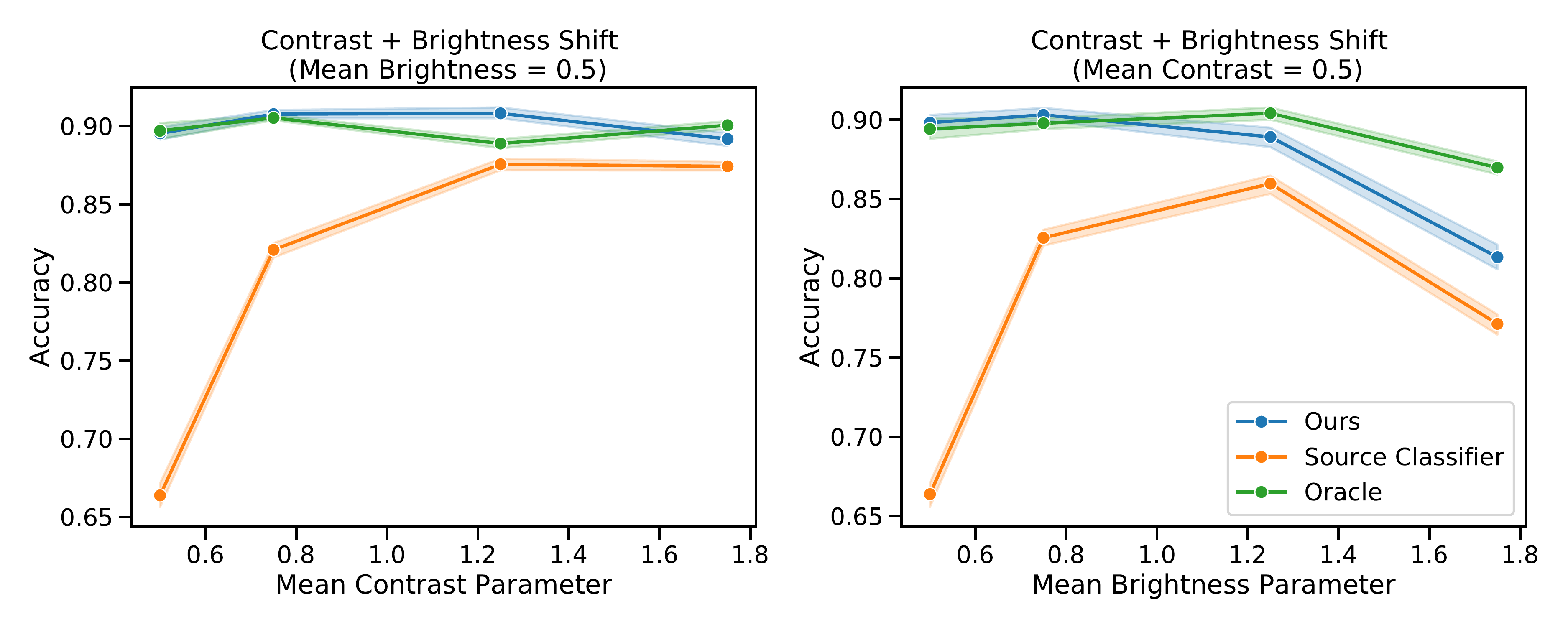}
    \caption{The target domain varies from the source along the axes of both brightness and contrast. In the first set of experiments, we construct the target domains by fixing the mean brightness of the target domains and sweeping over different mean contrasts. In the second set of experiments, we construct the target domains by fixing the mean contrast and sweeping over different mean brightness}
    \label{fig:multiple}
\end{figure*}

\textbf{Evaluation.} Let $D$ be the original CIFAR-10 test set. We construct the target domains as follows
\[\{\mathcal{D}^{t} \} = \{B_{\mu_{1}, 0.05}(C_{\mu_{2}, 0.05}(D)) \}. \]
In one set of experiments, we vary $\mu_{1} \in [0.5, 0.75, \dots 1.75]$ while setting $\mu_{2}=0.5$. In another set of experiments, we vary $\mu_{2} \in [0.5, 0.75, \dots 1.75]$ while setting $\mu_{1}=0.5$.

\textbf{Results.} We observe that our method recovers accuracy lost by the source classifier on the target domains (Figure \ref{fig:multiple}). Similar to Section \ref{sec:severity_range}, we see that when $\mu_{1} > 1$ our method offers an improvement over the source classifier but does not recover full accuracy. 
\end{subsubsection}
\end{subsection}

\subsection{MNIST Results}

\begin{subsubsection}{Comparison to Fine-Tuning}
\textbf{Baselines.} We compare our unsupervised method to the source classifier trained on the MNIST training set and oracle models as described in Section \ref{sec:setup}. The training data for each oracle model is generated by applying the corresponding forward transforms to the MNIST training set. Additionally, we compare to the fine-tuning methods described in Section \ref{sec:fine_tuning}. 

\textbf{Evaluation.} We consider mild and severe shifts along the axis of rotation. For the mild shift, the target domain is generated by applying the forward transforms $R_{30, 2}$ to the MNIST test set. For the severe shift, we apply $R_{60, 2}.$ Otherwise, the evaluation method is identical to Section \ref{sec:fine_tuning}.

\begin{figure*}[th]
\centering
    \includegraphics[width=0.5\textwidth]{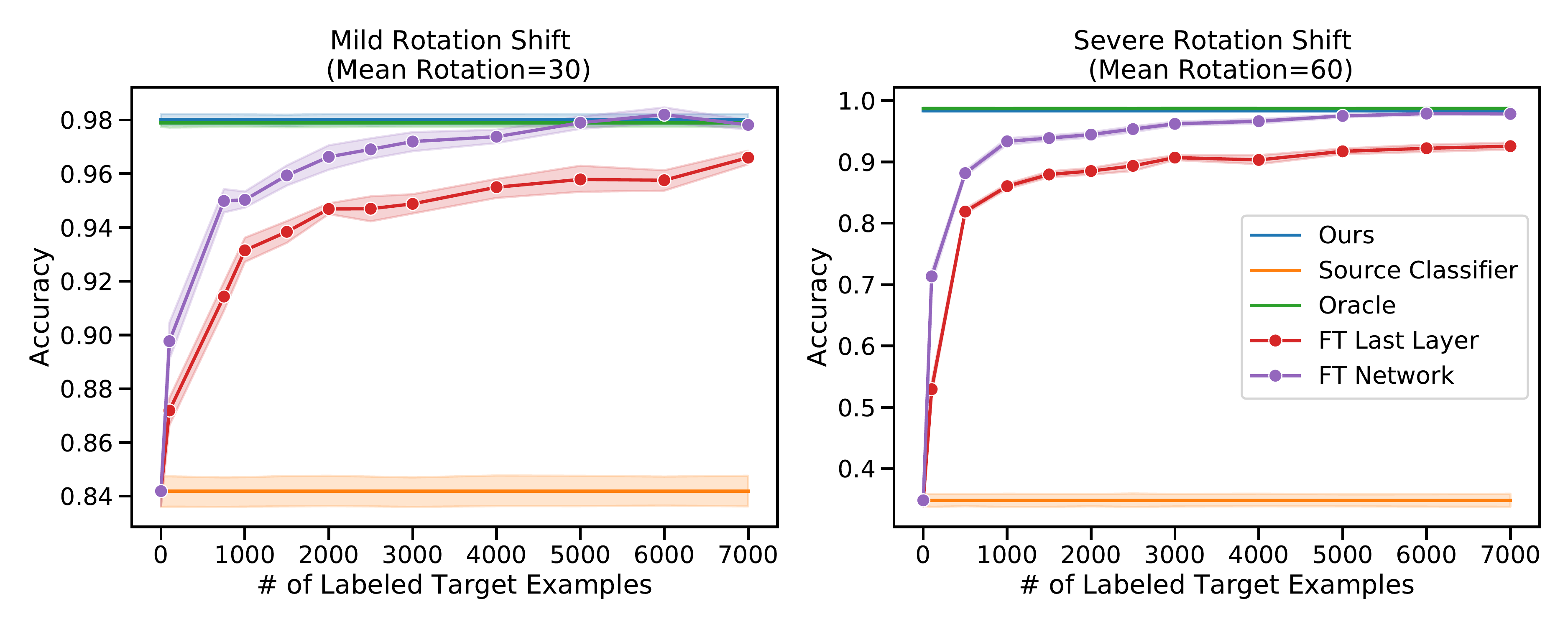}
    \caption{Accuracy achieved by fine-tuning a source classifier with $n$ labeled target examples. Fine-tuning the network (FT Network) requires at least $5000$ examples, and fine-tuning the last layer (FT Last Layer) does not achieve the same accuracy as our method even when all labeled examples are used. }
    
    \label{fig:tl_rot} 
\end{figure*}

\textbf{Results. } Across the mild and severe shifts, our unsupervised method excels compared to fine-tuning in the presence few labels. We outperform fine-tuning the final layer even when all $7000$ labeled target examples are available and fine-tuning the network when there are fewer than $5000$ target examples available (Figure \ref{fig:tl_rot}). Our method performs on par with the oracle models in these experiments. As in Section \ref{sec:fine_tuning}, the fine-tuning schemes and our method demonstrate accuracy improvements relative to the source classifier. 

\end{subsubsection}

\begin{subsubsection}{Effect of Shift Severity and Shift Range}
\textbf{Baselines.} As described in Section \ref{sec:setup}, we compare our method to 1) the source classifier trained on the MNIST training set and 2) oracle models, each trained on a target domain.

\textbf{Evaluation.} Let $D$ be the original MNIST test set. For the shift severity experiments, we construct the target domains as follows
\[ \{\mathcal{D}^{t}\} = \{R_{\mu, 2}(D) \text{ where } \mu \in [-60, 60] \}. \]
We limit $\mu \in [-60, 60]$ degrees because for large angles, the forward transform is not label-preserving. For the shift range experiments, we construct the target domains as
\[ \{\mathcal{D}^{t} \} = \{ R_{30, \sigma}(D) \text{ where } \sigma \in [2, 25] \}.\] 
We set $\mu=30$ in the shift range experiments to simulate a mild distribution shift (in contrast, with $\mu=0$, most of the generated examples are in-distribution for the source classifier).

\begin{figure*}[h]
    \centering
    \includegraphics[width=0.5\textwidth]{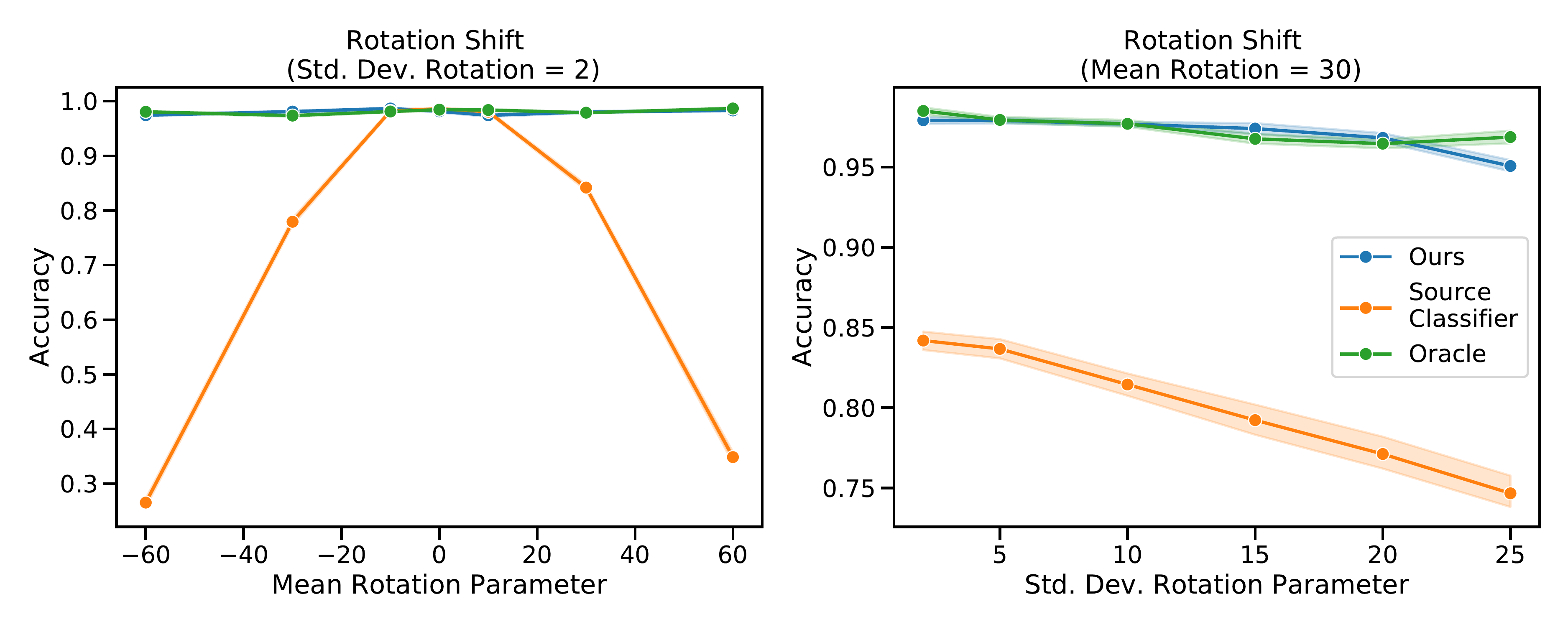}
   \caption{We evaluate how our method performs compared to the oracle model and the source classifier as the mean and standard deviation of the rotation shift changes.}
\label{fig:rot}
\end{figure*}

\textbf{Shift Severity Results.}  Our method recovers full accuracy, performing as well as the oracle model on these target domains (Figure \ref{fig:rot}). Our method is especially successful on this set of target domains because the rotation transform is invertible on MNIST.

\textbf{Shift Range Results.}  The source classifier's performance declines drastically as the shift range increases. In contrast, the performance of our method and that of the oracle models decline gradually (Figure \ref{fig:rot}).

\end{subsubsection}

\subsection{Experiment Details}

\subsubsection{Transformation Network Training} 
\label{sec:tnet}
The transformation network is a CNN. For CIFAR-10, we use the following architecture
\begin{enumerate} \itemsep0em
\item Convolutional layer with $3$ input channels, $6$ output channels, kernel size $5$, and stride $1$.
\item Maxpool layer with kernel size $2$ and stride $2$.
\item Convolutional layer with $6$ input channels, $16$ output channels, kernel size $5$, and stride $1$.
\item Linear layer with output size $120$.
\item Linear layer with output size $84$.
\item Linear layer with output size equal to number of transformation parameters.
\end{enumerate}

For MNIST, we modify the architecture slightly for single-channel images.

\begin{enumerate} \itemsep0em
\item Convolutional layer with $3$ input channels, $6$ output channels, kernel size $3$, and stride $1$.
\item Convolutional layer with $6$ input channels, $16$ output channels, kernel size $3$, and stride $1$.
\item Linear layer with output size $120$.
\item Linear layer with output size $84$.
\item Linear layer with output size equal to number of transformation parameters.
\end{enumerate}

We optimize the network weights using the Adam optimizer with learning rate 5e-5 and train for 30 epochs.

\subsubsection{Source Classifier Training}
We train each dataset's source classifier on the respective training set. The source classifier is trained on $80 \%$ and validated on $20 \%$ of the respective training set. Both source classifiers are trained until convergence. Classifiers are optimized using the Adam optimizer with learning rate 1e-3. For the CIFAR-10 experiments, the source classifier is a Resnet-18 model. In the MNIST experiments, the source classifier is a CNN with the following architecture: 
\begin{enumerate} \itemsep0em
\item Convolutional layer with $3$ input channels, $6$ output channels, kernel size $3$, and stride $1$.
\item Convolutional layer with $6$ input channels, $16$ output channels, kernel size $3$, and stride $1$.
\item Linear layer with output size $120$.
\item Linear layer with output size $84$.
\item Linear layer with output size equal to number of classes.
\end{enumerate}
When training on MNIST, no data augmentation is applied. When training on CIFAR-10, standard data augmentation is applied (horizontal flips, random crops).

\subsubsection{Oracle Model Training}
The oracle model is trained using the same procedure as the source classifier, except that the oracle model is trained on the target domain.  As described in Section \ref{sec:setup}, the training data for the oracle model is generated by applying forward transforms to the respective training dataset.

\end{document}